\documentclass[a4paper]{article}

\usepackage{INTERSPEECH2016}

\usepackage{graphicx}
\usepackage{amssymb,amsmath,bm}
\usepackage{textcomp}

\ninept

\title{Domain Adaptation of Recurrent Neural Networks for Natural Language Understanding}

\makeatletter
\def\name#1{\gdef\@name{#1\\}}
\makeatother \name{{\em Aaron Jaech$^1$, Larry Heck$^2$, Mari Ostendorf$^1$}}

\address{$^1$University of Washington \\
  $^2$Google Research \\
  {\small \tt ajaech@uw.edu, larryheck@google.com, ostendor@uw.edu}
}

\begin{document}

  \maketitle
  \begin{abstract}
    The goal of this paper is to use multi-task learning to efficiently scale
    slot filling models for natural language understanding to handle multiple
    target tasks or domains. The key to scalability is reducing the amount of
    training data needed to learn a model for a new task. The proposed
    multi-task model delivers better performance with less data by leveraging
    patterns that it learns from the other tasks. The approach supports an open
    vocabulary, which allows the models to generalize to unseen words,
    which is particularly important when very little training data is used.
    A newly collected crowd-sourced data set, covering four different domains,
    is used to demonstrate the effectiveness of the domain adaptation and 
    open vocabulary techniques.
    
  \end{abstract}

\section{Introduction}

Slot filling models are a useful method for simple natural language understanding tasks, where
information can be extracted from a sentence and used to perform some structured action. For example, dates, departure cities and destinations represent slots to fill in a flight booking task. This information is extracted from natural language queries leveraging typical context associated with each slot type. Researchers have been exploring data-driven approaches to learning models for automatic identification of slot information since the 90's, and significant advances have been made \cite{price1990evaluation}. Our paper builds on recent work on slot-filling using recurrent neural networks (RNNs) with a focus on the problem of training from minimal annotated data, taking an approach of sharing data from multiple tasks to reduce the amount of data for developing a new task.

As candidate tasks, we consider the actions that a user might perform via apps on their phone. 
Typically, a separate slot-filling model would be trained for each app.
For example, one model understands queries about classified ads for cars \cite{meng1996wheels} 
and another model handles queries about the weather \cite{glass1998telephone}.
As the number of apps increases, this approach becomes impractical due to the
burden of collecting and labeling the training data for each model. In addition, using independent models for 
each task has high storage costs for mobile devices.

Alternatively, a single model can be learned to handle all of the apps. This type of approach is known as 
multi-task learning and can lead to improved performance on all of the tasks due to information sharing
between the different apps \cite{caruana1997multitask}. Multi-task learning in combination with neural 
networks has been shown to be effective for natural language processing tasks \cite{collobert2008unified}.
When using RNNs for slot filling, almost all of the model parameters can be
shared between tasks. In our study, only the relatively small output layer, which consists of slot embeddings, is individual to each app.
More sharing means that less training data per app can be used and there will still be enough data to effectively train the network.
The multi-task approach has lower data requirements, which leads to a large cost
savings and makes this approach scalable to large numbers of applications. 

The shared representation that we build on leverages recent work 
on slot filling models that use neural network based approaches.
Early neural network based papers propose feedforward \cite{deoras2013deep} or RNN architectures \cite{mesnil2013investigation,yao2013recurrent}. The focus shifted to RNN's with long-short term memory cells (LSTMs) \cite{xu2013convolutional,yao2014spoken,shi2015contextual,mesnil2015using}
after LSTMs were shown to be effective for other tasks \cite{sundermeyer2012lstm}.
The most recent papers use variations on LSTM sequence models, including encoder-decoder, external memory, or attention architectures \cite{peng2015recurrent,kurata2016leveraging,dong2016language}.  The particular variant that we build on is a bidirectional LSTM, similar to \cite{newatis,mesnil2015using}.

One highly desirable property of a good slot filling model is to generalize to previously
unseen slot values. For instance, we should not expect that the model will see the
names of all the cities during training time, especially when only a small
amount of training data is used. We address the generalizability
issue by incorporating the open vocabulary embeddings from Ling et al. 
into our model \cite{ling2015finding}. These embeddings work by using a character RNN to process a word one letter at a time.
This way the model can learn to share parameters between different words that use the same morphemes.
For example BBQ restaurants frequently use words like ``smokehouse'', ``steakhouse'', and ``roadhouse'' in their names 
and ``Bayside'',``Bayview'', and ``Baywood'' are all streets in San Francisco. Recognizing these patterns would 
be helpful in detecting a restaurant or street name slot, respectively.

The two main contributions of this work are the multi-task model and the use of the open vocabulary character-based embeddings, which together allow for scalable slot filling models.
Our work on multi-task learning in slot filling differs from its previous use in \cite{li2011multi}
in that we allow for soft sharing between tasks instead of explicitly matching slots to
each other across different tasks.  A limitation of explicit slot matching is that two slots that appear to have the same underlying type, such as location-based slots, may actually use the slot information in different ways depending on the overall intent of the 
task. In our model, the sharing between tasks is done implicitly by the neural network. 
Our approach to handling words unseen in training data is different from the delexicalization proposed in \cite{Henderson+14} in that we do not require the vocabulary items associated with slots and values to be prespecified. It is complementary to work on extending domain coverage  \cite{Gasic+13,ElKahky+14}.

The proposed model is described in more detail in Section~\ref{sec:model}.
The approach is assessed on a new data collection based on four apps, described in Section~\ref{sec:data}.  The experiments described in Section~\ref{sec:expts} investigate how much data is necessary for the $n$-th app using a multi-task model that leverages the data from the previous $n-1$ apps, with results compared against the single-task model that only utilizes the data from the $n$-th app. We conclude in Section~\ref{sec:concl} with a summary of the key findings and discussion of opportunities for future work.

\section{Model}
 \label{sec:model}

Our model has a word embedding layer, followed by a bi-directional LSTM (bi-LSTM), and a softmax output layer.
The bi-LSTM allows the model to use information from both the right and left contexts of 
each word when making predictions. We choose this architecture because similar models have been used 
in prior work on slot filling and have achieved good results \cite{newatis,mesnil2015using}.
The LSTM gates are used as defined by Sak et al.\ including the use of the linear projection layer on
the output of the LSTM \cite{sak2014long}. The purpose of the projection layer is to produce a model
with fewer parameters without reducing the number of LSTM memory cells. 
For the multi-task model, the word embeddings and the bi-LSTM parameters are shared across tasks but each task
has its own softmax layer. This means that if the multi-task model has half a million parameters, only a couple
thousand of them are unique to each task and the other 99.5\% are shared between all of the tasks.

The slot labels are encoded in BIO format \cite{ramshaw1999text} indicating if a word is the
beginning, inside or outside any particular slot. Decoding is done greedily. 
If a label does not follow the BIO syntax rules, i.e. an inside tag 
must follow the appropriate begin tag, then it is replaced with the outside label.
Evaluation is done using the CoNLL evaluation script \cite{tjong2000introduction} to
calculate the F1 score. This is the standard way of evaluating slot-filling models
in the literature.

In recent work on language modeling, a neural architecture that combined fixed word embeddings
with character-based embeddings was found to 
to be useful for handling previously unseen words
\cite{jozefowicz2016exploring}.
Based on that result, the embeddings in the open vocabulary model are a concatenation of the
character-based embeddings with fixed word embeddings. When an out-of-vocabulary word
is encountered, its character-based embedding is concatenated with the embedding for the unknown
word token.
The character-based embeddings are generated from a two layer bi-LSTM that processes
each word one character at a time. The character-based word embedding is produced
by concatenating the last states from each of the directional LSTM's in the second layer and 
passing them through a linear layer for dimensionality reduction.

\section{Data}
\label{sec:data}

Crowd-sourced data was collected simulating common use cases for four different apps:
United Airlines, Airbnb, Greyhound bus service and OpenTable. The corresponding actions
are booking a flight, renting a home, buying bus tickets, and making a reservation at a restaurant.
In order to elicit natural language, crowd workers were instructed to simulate a conversation
with a friend planning an activity as opposed to giving a command to the computer. 
Workers were prompted with a slot type/value pair and asked to form a 
reply to their friend using that information. The instructions were to not include any other
potential slots in the sentence but this instruction was not always followed by the workers.

\begin{table}[]
\centering
\begin{tabular}{crr}
\textbf{Data set} & \textbf{Queries} & \multicolumn{1}{l}{\textbf{Slot Types}} \\ \hline
United App & 20,697 & 12 \\
OpenTable & 3,151 & 6 \\
Greyhound & 4,951 & 13 \\
Airbnb & 4,666 & 11
\end{tabular}
\caption{Data statistics for each of the four target applications.}
\label{table:dataset_stats}
\end{table}

\begin{table}[]
\centering
\begin{tabular}{cp{5.7cm}}
\textbf{App} & \multicolumn{1}{c}{\textbf{Slot Types}} \\ \hline
Airbnb & number of people, type of room, desired amenities, start and end dates, date range, location, listing type and three price-related slots (desired price, lower and upper bounds)  \\
Greyhound & date and time for the departure and return, departure and return locations, number of children, adults, and seniors, promotion code, discount type, whether the trip is one way, and wheelchair use \\
OpenTable & cuisine, date, time, location, number of people, and restaurant name \\
United &  return \& departure dates and locations ($\times 2$ for multi-hop), ticket quantity, $\pm$ nonstop, ticket class, and whether or not the flight is one way or round trip or multi-hop
\end{tabular}
\caption{Listing of slot types for each app.}
\label{table:slottypes}
\end{table}

Slot types were chosen to roughly correspond to form fields and UI elements, such as check boxes or dropdown menus,
on the respective apps. The amount of data collected per app and the number of slot types is listed in Table \ref{table:dataset_stats}.
The slot types for each app are described in Table \ref{table:slottypes}, and an example labeled sentence from each app is given in Table \ref{table:exsentences}.
One thing to notice is that the the number of slot types is relatively small when compared to the popular ATIS dataset that has
over one hundred slot types \cite{price1990evaluation}. In ATIS, separate slot types would be used for names of cities, states, or countries whereas in this data all of those would fall under a single slot for locations.

Slot values were pulled from manually created lists of locations, dates and times, restaurants, etc.
Values for prompting each rater were sampled from these lists. Workers were instructed to
use different re-phrasings of the prompted values, but most people used the prompted
value verbatim. Occasionally, workers used an unprompted slot value not in the list.

\begin{table*}[t]
\centering
\begin{tabular}{cc} 
\textbf{App} & \textbf{Example Sentence} \\ \hline
Airbnb & I want to keep the price below \textless PriceUpper\textgreater { }\$1300 per week \textless /PriceUpper\textgreater . \\
Greyhound & We should return on \textless ReturnDate\textgreater { }Jan 11 \textless /ReturnDate\textgreater \\
OpenTable & Let's do something on \textless Loc\textgreater { }Castro Street \textless /Loc\textgreater \\
United & please book flight from \textless FromLoc\textgreater { }burbank \textless /FromLoc\textgreater { }to \textless ToLoc\textgreater { }st petersburg \textless /ToLoc\textgreater \\ 
\end{tabular}
\label{table:exsentences}
\caption{Example labeled sentences from each application.}
\end{table*}

For the word-level LSTM, the data was lower-cased and tokenized using a standard tokenizer. Spelling mistakes were not corrected.
All digits were replaced by the '\#' character. Words that appear only once in the training data are replaced with
an unknown word token. For the character-based word embeddings used in the open vocabulary model, no lower casing or digit replacement is done.

Due to the way the OpenTable data was collected some slot values were over-represented leading 
to over fitting to those particular values. To correct this problem sentences that used the over-represented 
slot values had their values replaced by sampling from a larger list of potential values. The affected slot types
are the ones for cuisine, restaurant names, and locations. This substitution made the OpenTable data 
more realistic as well as more similar to the other data that was collected.

The data we collected for the United Airlines app is an exception in a few ways: we collected four times as much
data for this app than the other ones; workers were occasionally prompted with up to
four slot type/value pairs; and workers were instructed to give commands to their device
instead of simulating a conversation with a friend.  For all of the other apps, workers were prompted to use a single
slot type per sentence. 
We argue that having varying amounts of data for different apps is a realistic scenario.

Another possible source of data is the Air Travel Information Service
(ATIS) data set collected in the early 1990's \cite{price1990evaluation}. However, this data is sufficiently similar to the United collection, that it is not likely to add sufficient variety to improve the target domains. Further, it suffers from artifacts of data collected at a time with speech recognition systems had much higher error rates. The new data collected for this work fills a need raised in  \cite{tur2010left}, which concluded that lack of data
was an impediment to progress in slot filling.

\section{Experiments}
\label{sec:expts}

The section describes two sets of experiments: the first is designed to test the effectiveness
of the multi-task model and the second is designed to test the generalizability of the
open vocabulary model.
The scenario is that we already have $n-1$ models in place and we wish to discover how
much data will be necessary to build a model for an additional application.

\subsection{Training and Model Configuration Details}

The data is split to use 30\% for training with 70\% to be used for test data. The reason that a majority
of the data is used for testing is that in the second experiment the results are reported separately for
sentences containing out of vocabulary tokens and a large amount of data is needed to get a sufficient
sample size. Hyperparameter tuning presents a challenge when operating in a low resource scenario.
When there is barely enough data to train the model none can be spared for a validation set. We used
data from the United app for hyperparameter tuning since it is the largest and assumed that the hyperparameter
settings generalized to the other apps.

Training is done using stochastic gradient descent with minibatches of 25 sentences. 
The initial learning rate is 0.3 and is set to decay to 98\% of its value every 100 minibatches.
For the multi-task model, training proceeds by alternating between each of the tasks when selecting the
next minibatch.
All the parameters are initialized uniformly in the range [-0.1, 0.1].
Dropout is used for regularization
on the word embeddings and on the outputs from each LSTM layer with the dropout probability set to 60\% \cite{zaremba2014recurrent}.

For the single-task model, the word embeddings are 60 dimensional and the LSTM is dimension 100 with a 70 dimensional projection layer on the LSTM.
For the multi-task model, word embeddings are 200 dimensional, and the LSTM has 250 dimensions with a 170 dimensional projection layer.
For the open vocabulary version of the model, the 200-dimensional input is a concatenation of 160-dimensional traditional word embeddings with 40-dimensional character-based word embeddings. 
The character embedding layer is 15 dimensions, the first LSTM layer is 40 dimensions with a 20 dimensional projection layer, and the second LSTM layer is 130 dimensions. 

\subsection{Multi-task Model Experiments}

We compare a single-task model against the multi-task model for
varying amounts of training data. In the multi-task model, the full amount of data is used
for $n-1$ apps and the amount of data is allowed to vary only for the $n$-th application.
These experiments use the traditional word embeddings with a closed vocabulary.
Since the data for the United app is bigger than the other three apps combined, it is used as an
anchor for the multi-task model. The other three apps alternate in the position of the $n$-th app.
The data usage for the $n$-th app is varied while the other $n-1$ apps in each experiment use the full amount of available
training data. The full amount of training data is different for each app.
The data used for the $n$-th app is 200, 400, or 800 sentences or all available training data depending on the experiment.
The test set remains fixed for all of the experiments even as part of the training data is discarded to simulate the low resource scenario.

\begin{figure}[h]
 \centering
 \includegraphics[width=0.45\textwidth]{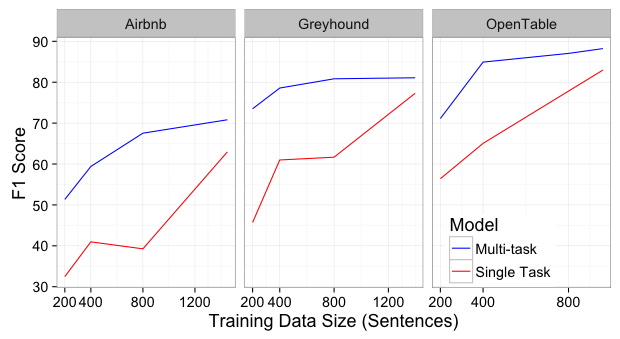}
 \caption{F1 score for multi-task vs.\ single-task models.}
 \label{fig:expresults}
 \end{figure}

In Figure \ref{fig:expresults} we show the single-task vs.\ multi-task model performance 
for each of three different applications. The multi-task model outperforms
the single-task model at all data sizes, and the relative performance increases
as the size of the training data decreases. When only 200 sentences of training data are used,
the performance of the multi-task model is about 60\% better than the single-task model for both
the Airbnb and Greyhound apps. The relative gain for the OpenTable app is 26\%.
Because the performance of the multi-task model decays much more slowly as the amount of training data is reduced, 
the multi-task model can deliver the same performance with a considerable reduction in the amount of labeled data.

\subsection{Open Vocabulary Model Experiments}

The open vocabulary model experiments test the ability of the model to handle unseen
words in test time, which are particularly likely to occur when using a reduced amount of
training data. In these experiments the open vocabulary model is compared against
the fixed embedding model. The results are reported separately for the sentences that 
contain out of vocabulary tokens, since these are where the open vocabulary system
is expected to have an advantage.

\begin{figure}[h]
\centering
\includegraphics[width=0.4\textwidth]{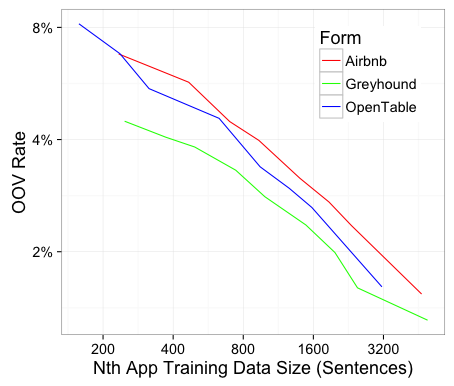}
\caption{OOV rate for each of the n apps.}
\label{fig:oovs}
\end{figure}

Figure \ref{fig:oovs} gives the OOV rate for each app for varying amounts of training data plotted on a log-log scale.
The OOV words tend to be task-specific terminology. For example, the OpenTable task is the only
one that has names of restaurants but names of cities are present in all four tasks so they tend to be covered
better. The OOV rate dramatically increases when the size of the training data is less than
500 sentences. Since our goal is to operate in the regime of less than 500 sentences per task,
handling OOVs is a priority.
%
The  multi-task model is used in these experiments.
The only difference between the closed vocabulary and open vocabulary systems is that the closed
vocabulary system uses the traditional
word embeddings and the open vocabulary system uses
the traditional word embeddings concatenated with character-based embeddings.

\begin{table}[h]
\centering
\begin{tabular}{ccccc}
 & \multicolumn{2}{c}{\textbf{Full Test Set}} &\multicolumn{2}{c}{\textbf{OOV Subset}} \\ 
\textbf{Vocabulary} & \textbf{Closed} & \textbf{Open} & \textbf{Closed} & \textbf{Open}\\  \hline
Airbnb     & 74.4 & 72.7 & 54.5 & 58.2       \\
Greyhound  &  85.2 & 84.4 &  64.2 & 67.0     \\
OpenTable  & 89.7& 88.9 &  68.8 & 68.1   \\
United     & 90.8 & 90.6  & 81.8 & 80.7 \\
\end{tabular}
\caption{Comparison of F1 scores for open and closed vocabulary systems on the full test set vs.\ the subset 
with OOV words.}
\label{table:openvocabresults}
\end{table}

Table \ref{table:openvocabresults} reports F1 scores on the test set for 
both the closed and open vocabulary systems. The results differ between
the tasks, but none have an overall benefit from the open vocabulary system.  
Looking at the subset of
sentences that contain an OOV token, the open vocabulary system
delivers increased performance on the Airbnb and Greyhound tasks. 
These two are the most difficult apps out of the four and therefore
had the most room for improvement. 
The United app is also all lower case and casing is an important clue for detecting proper nouns that the
open vocabulary model takes advantage of.

\begin{figure}[h]
\centering
\includegraphics[width=0.46\textwidth]{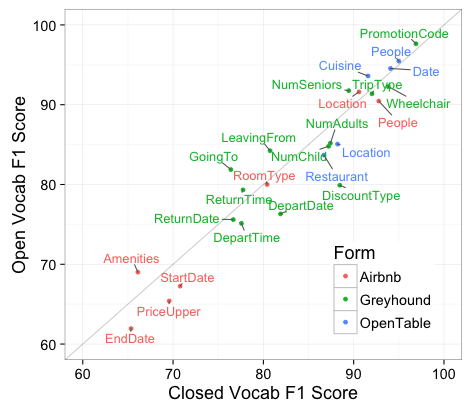}
\caption{Comparison of performance on individual slot types.}
\label{fig:fscores}
\end{figure}

Looking a little deeper, in Figure \ref{fig:fscores} we show the breakdown in performance
across individual slot types. Only those slot types which occur at least one hundred times in the test
data are shown in this figure. The slot types that are above the diagonal saw a performance
improvement using the open vocabulary model. The opposite is true for those that are 
below the diagonal. The open vocabulary 
system appears to do worse on slots that express quantities, dates and times and better on slots with
greater slot perplexity (i.e., greater variation in slot values) like ones relating to locations. The three slots where the open
vocabulary model gave the biggest gain are the Greyhound \textit{LeavingFrom} and \textit{GoingTo}
slots along with the Airbnb \textit{Amenities} slot. The three slots where the open vocabulary model
did the worst relative to the closed vocabulary model are the Airbnb \textit{Price} slot, along with the
Greyhound \textit{DiscountType} and \textit{DepartDate} slots. The \textit{Amenities} slot is an example of a slot with higher perplexity (with options related to pets, availability of a gym, parking, fire extinguishers, proximity to attractions),
and the \textit{DiscountType} is one with lower perplexity (three options cover almost all cases). We hypothesize that the reason that the numerical slots are better under the
closed vocabulary model is due to their relative simplicity and not an inability of the character
embeddings to learn representations for numbers.

\section{Conclusions}
\label{sec:concl}

In summary, we find that using a multi-task model with shared embeddings gives a large reduction in the minimum amount of data needed
to train a slot-filling model for a new app. This translates into a cost savings for deploying slot filling models
for new applications. The combination of the multi-task model with the open vocabulary embeddings increases
the generalizability of the model especially when there are OOVs in the sentence.
These contributions allow for scalable slot filling models.

For future work, there are some improvements that could be made to the model such as the addition of an
attentional mechanism to help with long distance dependencies \cite{dong2016language}, 
use of beam-search to improve decoding, and exploring unsupervised adaptation as in \cite{Henderson+14}.

Another item for future work is to collect additional tasks to examine the scalability
of the multi-task model beyond the four applications that were used in this work. 
Due to their extra depth, character-based methods usually require more data than
word based models \cite{srivastava2015highway}.
Since this paper uses limited data, the collection of additional tasks may
significantly improve the performance of the open vocabulary model.

  \eightpt
  \bibliographystyle{IEEEtran}

  \bibliography{mybib}

\end{document}